\documentclass[5pt]{article}

\usepackage[letterpaper]{geometry}
\usepackage{spconf,amsmath,epsfig}

\usepackage{times}
\usepackage{epsfig}
\usepackage{graphicx}
\usepackage{tabularx}
\usepackage{enumerate}

\usepackage{amsmath,amssymb} 

\usepackage{amsfonts}            

\usepackage{bm}

\usepackage{color}

\usepackage{url}

\usepackage{algorithm}               
\usepackage{algorithmic}            
\usepackage{multirow}                
\usepackage{xcolor} 

\usepackage{fancyhdr}
\begin{document}\sloppy

\def\x{{\mathbf x}}
\def\L{{\cal L}}

\title{Efficient Privacy Preserving Viola-Jones Type Object Detection \\ via Random Base Image Representation}
%
\name{Xin Jin$^{1}$, Peng Yuan$^{1,2}$, Xiaodong Li$^{1,*}$, Chenggen Song$^{1}$, Shiming Ge$^{3}$, Geng Zhao$^{1}$, Yingya Chen$^{1}$ \thanks{This work is partially supported by the National Natural Science Foundation of China (Grant NO.61402021, 61402023, 61640216), the Science and Technology Project of the State Archives Administrator (Grant NO. 2015-B-10), the open funding project of State Key Laboratory of Virtual Reality Technology and Systems, Beihang University (Grant NO. BUAA-VR-16KF-09), and the Fundamental Research Funds for the Central Universities (NO. 2016LG03, 2016LG04).}}

\address{$^{1}$Beijing Electronic Science and Technology Institute,
Beijing 100070, China\\
$^{2}$Xidian University, Xi'an 710071, China\\
$^{3}$Institute of Information Engineering, Chinese Academy of Sciences, Beijing 100095, China\\
*Corresponding Author: lxd@besti.edu.cn}

%

\maketitle

\begin{abstract}
A cloud server spent a lot of time, energy and money to train a Viola-Jones type object detector \cite{Viola2001} with high accuracy. Clients can upload their photos to the cloud server to find objects. However, the client does not want the leakage of the content of his/her photos. In the meanwhile, the cloud server is also reluctant to leak any parameters of the trained object detectors. 10 years ago, Avidan \& Butman introduced \emph{Blind Vision}, which is a method for securely evaluating a Viola-Jones type object detector. Blind Vision uses standard cryptographic tools and is painfully slow to compute, taking a couple of hours to scan a single image. The purpose of this work is to explore an efficient method that can speed up the process. We propose the \emph{Random Base Image (RBI) Representation}. The original image is divided into random base images. Only the base images are submitted randomly to the cloud server. Thus, the content of the image can not be leaked. In the meanwhile, a random vector and the secure Millionaire protocol are leveraged to protect the parameters of the trained object detector. The RBI makes the integral-image enable again for the great acceleration. The experimental results reveal that our method can retain the detection accuracy of that of the plain vision algorithm and is significantly faster than the traditional blind vision, with only a very low probability of the information leakage theoretically.
\end{abstract}
\begin{keywords}
Blind Vision, Random Base Image, Privacy Preserving, Object Detection
\end{keywords}

\section{Introduction}

Recently, widespread smart phones with cameras enable people to shot images and videos nearly anytime and anywhere. Millions of surveillance cameras including the driving recorders captures images and videos every second. All these conveniences devices are producing the large-scale visual media data, which is considered as the \emph{biggest big data}.

\begin{figure}
\centering
\includegraphics[height=3.4cm]{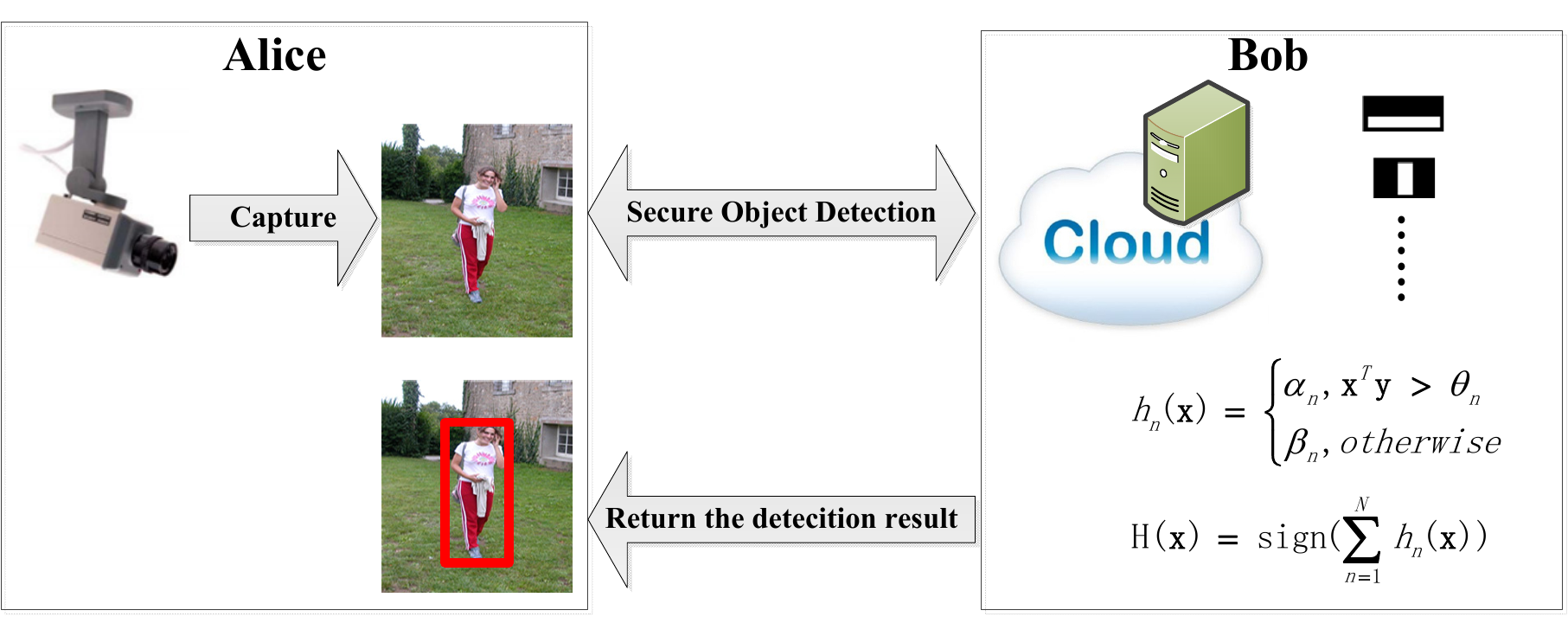}
\caption{Alice would like to detect objects in a collection of sensitive surveillance images she own. Bob has an object detection algorithm that he is willing to let Alice use, for a fee, as long as she learns nothing about his detector. Alice is willing to use Bob's detector provided that he will learn nothing about her images, not even the result of the object detection operation \cite{Avidan2006a}.}
\label{fig:scenario}
\end{figure}

Due to the limited storage space of these terminal devices, large-scale visual media data is being uploaded and stored in the cloud servers. Not only the storage, but also the processing of large-scale visual media data are being outsourced to the cloud servers. 

The cloud servers have some strong algorithms such as face/object detection, face/object recognition, intelligent video surveillance. Nowadays, people can easily find all the faces in their photos stored in the cloud servers using the powerful face detection algorithms maintained by the cloud servers. However, the cloud servers are always third party entities. Thus the privacy of the users' visual media data may be leaked to the public or unauthorized parties.

In the meanwhile, the powerful cloud services for visual media analysis and processing need a lot of money, data and time from the cloud server producers. The cloud servers are also reluctant to leak any parameters of the trained models or some protected details of their algorithms with copyrights.

Thus, the privacy of both the content of the visual media from the clients and the parameters of the vision algorithms from the cloud servers should be protected. 10 years ago, Avidan \& Butman introduced \emph{Blind Vision} \cite{Avidan2006a}, which is a method for securely evaluating a Viola-Jones type face detector. Blind Vision uses standard cryptographic tools and is painfully slow to compute, taking a couple of hours to scan a single image. 

After that, rich literatures have been proposed in this field. The cryptographic tools such as secret sharing (SS) \cite{Upmanyu2009}, security multi-party computation (SMC) \cite{Osadchy2010}, homomorphic encryption (HE) \cite{Sohn2010}, garbed circuit (GC) \cite{Chun2014}, Chaotic System (CS) \cite{Jin2016} are heavily used. Plenty of computer vision applications have been modified to the privacy preserving or secure versions such as private face detection \cite{Avidan2006b}, face recognition \cite{Osadchy2010}, content based image retrieval \cite{Shashank2008}, visual media search on public datasets \cite{Fanti2013}, intelligent video surveillance \cite{Upmanyu2009,Sohn2010,Chun2014,Jin2016}.

However, most of these work rely heavily on cryptographic tools, which are painfully slow to compute or need bit by bit interaction between the clients and the cloud servers. In this paper, we revisit the \emph{Blind Vision} \cite{Avidan2006a} and attempt to make the blind vision towards \emph{cryptographic-free}, without losing the security properties. We use  randomness and only a little cryptographic operations to protect the visual media data of the clients and the parameters of the trained models in the cloud servers.

A novel image representation called \emph{Random Base Image(RBI)} representation is proposed. In this work, we also investigate the object detection in the cloud. We apply our RBI to the famous Viola and Jones object detection method and propose a novel blind object detection method.  We separate an image into random base images. The weight of each base image is only known by the client. The base images are sent randomly to the cloud server. The cloud server cannot recover anything from the random base images. A random vector  and the secure Millionaire protocol \cite{Avidan2006a} are leveraged to protect the parameters of the trained object detector. The RBI makes the integral-image enable again for the great acceleration. The experimental results reveal that our method is significantly faster than the traditional blind vision, with only a very low probability of the information leakage theoretically.

\begin{figure*}
\centering
\includegraphics[height=6.5cm]{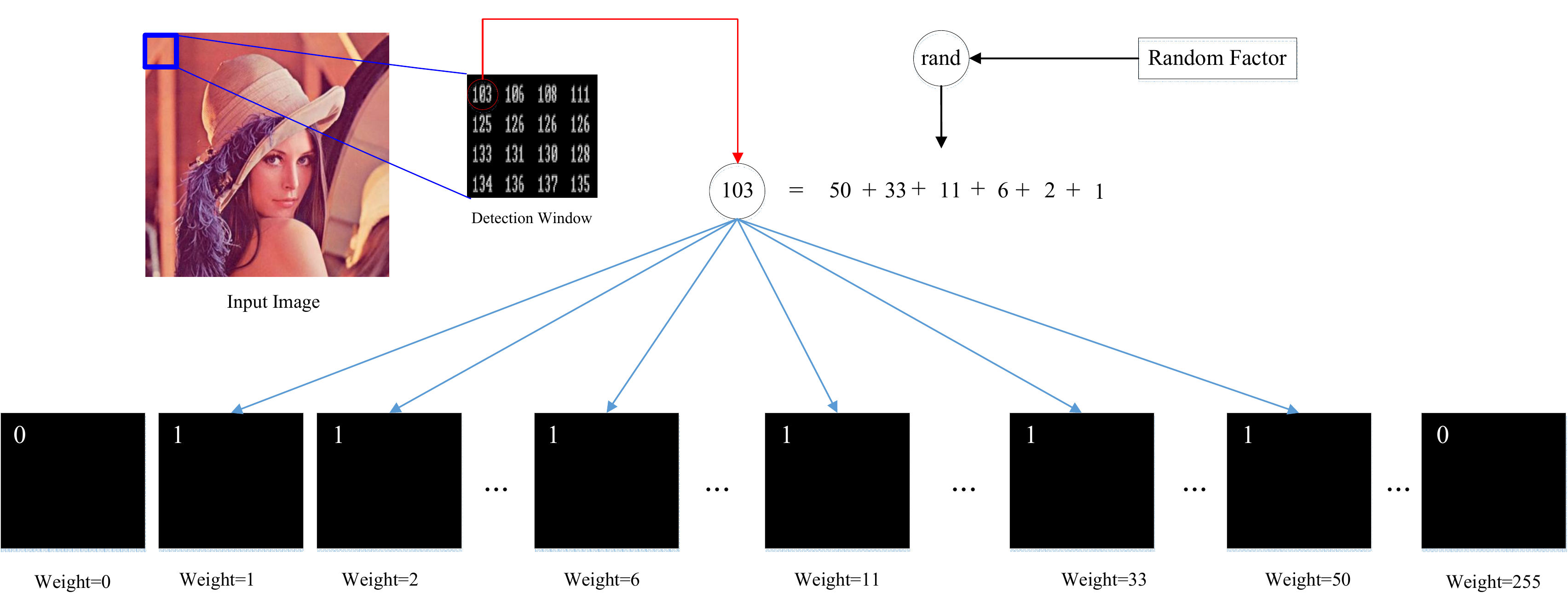}
\caption{The proposed Random Base Image Representation. For each single pixel $p_x$ in the detection window $x$ ($p_x \in [0,255]$), we randomly select $S$ numbers $p_x^{j}, j=1,2,...,S, p_x^{j} \in [0,p_x]$ to represent $p_x = \sum_{j=1}^{S}p_x^{j}$. We fix the number of the base image to $M=256$ and set each weight $w_i = i$. The initial value of each pixel in each base image is set to $0$. Then we set the corresponding pixel in each of the ${p_x^{j}}^{th}$ base image $B_i$ to $1$: $p_{B_i}=1, i=p_x^{j}$.}
\label{fig:RBI}
\end{figure*}

\section{Secure Object Detection}

In this section we develop a secure object detector with the random base image representation.

\subsection{Notations}

Our scenario and the notations are the same as that of traditional Blind Vision \cite{Avidan2006a}, as show in Figure \ref{fig:scenario}. Denote some $L$ dimensions finite field $F$ that is large enough to represent all the intermediate results. Denote by $X$ the image that Alice owns. A particular detection window within the image $X$ will be denoted by $x \in F^L$ and $x$ will be treated in vector form. Bob owns a strong classifier of the form

\begin{eqnarray}
H(\mathbf{x}) = \operatorname{sign}(\sum_{n=1}^{N}h_n(\mathbf{x})),
\label{eq:strong}
\end{eqnarray}
where $h_n(\mathbf{x})$ is a threshold function of the form

\begin{eqnarray}
h_n(\mathbf{x}) = 
\begin{cases}
 & \alpha_n ~~  \mathbf{x}^{T}\mathbf{y_n} > \theta_n \\ 
 & \beta_n  ~~ \text{otherwise,}  
\end{cases}
\label{eq:weak}
\end{eqnarray}
and $y_n \in F^L$ is the hyperplane of the threshold function $h_n(\mathbf{x})$. The parameters $\alpha_n \in F, \beta_n \in F$ and $\theta_n \in F$ of $h_n(\mathbf{x})$ are determined during training; $N$ is the number of weak classifiers used.

\subsection{The Random Base Image Representation}
The core idea of our RBI is to separate the original image into some random base images with fixed weights. The original image can be recovered by all the base images. The sparse representation can be considered as the one has such ability. However, they need another image dataset for learning the base images. Further more, there could be reconstruction error. Thus, we fix the weights and randomize the base images themselves.

The detection window $\mathbf{x}$ can be represented as:

\begin{eqnarray}
\mathbf{x} = \sum_{i=0}^{M-1}w_i\mathbf{B}_i,
\label{eq:linear}
\end{eqnarray}
where $\mathbf{B}_i$ is the base image with weight $w_i$. As is shown in Figure \ref{fig:RBI}, each base image has a fixed weight. The base image itself is randomly determined. The number of the base image is set to $M=256$. Thus, each base image can be a binary image, which is easy for network transfer and fast to compute. In addition, there are $256!$ permutation of the base image which is not easy to guess. The process of the RBI generation is described in Algorithm \ref{alg:RBI}.

\begin{algorithm}[htb]         
\caption{Random Base Image Factorization}             
\label{alg:RBI}                  
\begin{algorithmic}[1]                
\REQUIRE ~~\\                       
    The detection window $\mathbf{x}$ from the client image $\mathbf{X}$
\ENSURE ~~\\                         
   $M$ random binary base images\\
    $\mathbf{B}=\{\mathbf{B}_0, \mathbf{B}_1,..., \mathbf{B}_{M-1}\}$
\STATE Alice creates $M$ binary images with all the pixels initialized to $0$. Each binary image has the same size as that of the detection window $\mathbf{x}$.

\STATE The weight of each binary image is set to $w_i=i, i=0,1,...,M-1$.


\STATE Alice sets a index $j=1$. For each pixel $p_x$ in $x$, Alice repeat the following 3 steps until $p_x=0$.

(1) Alice generates a random number $p_x^{j} \in [0, p_x]$.\\ 
(2) Set the corresponding pixel in the ${p_x^{j}}^{th}$ base image $B_i$ to $1$: $p_{B_i}=1, i=p_x^{j}$.\\
(3) $p_x = p_x - p_x^{j}, j = j+1$.

\RETURN $\mathbf{B}=\{\mathbf{B}_0, \mathbf{B}_1,..., \mathbf{B}_{M-1}\}$.               
\end{algorithmic} 
\end{algorithm} 

\subsection{Secure Object Detection with RBI}

\subsubsection{Secure Object Classifier Protocol}
\begin{figure*}
\centering
\includegraphics[height=8.3cm]{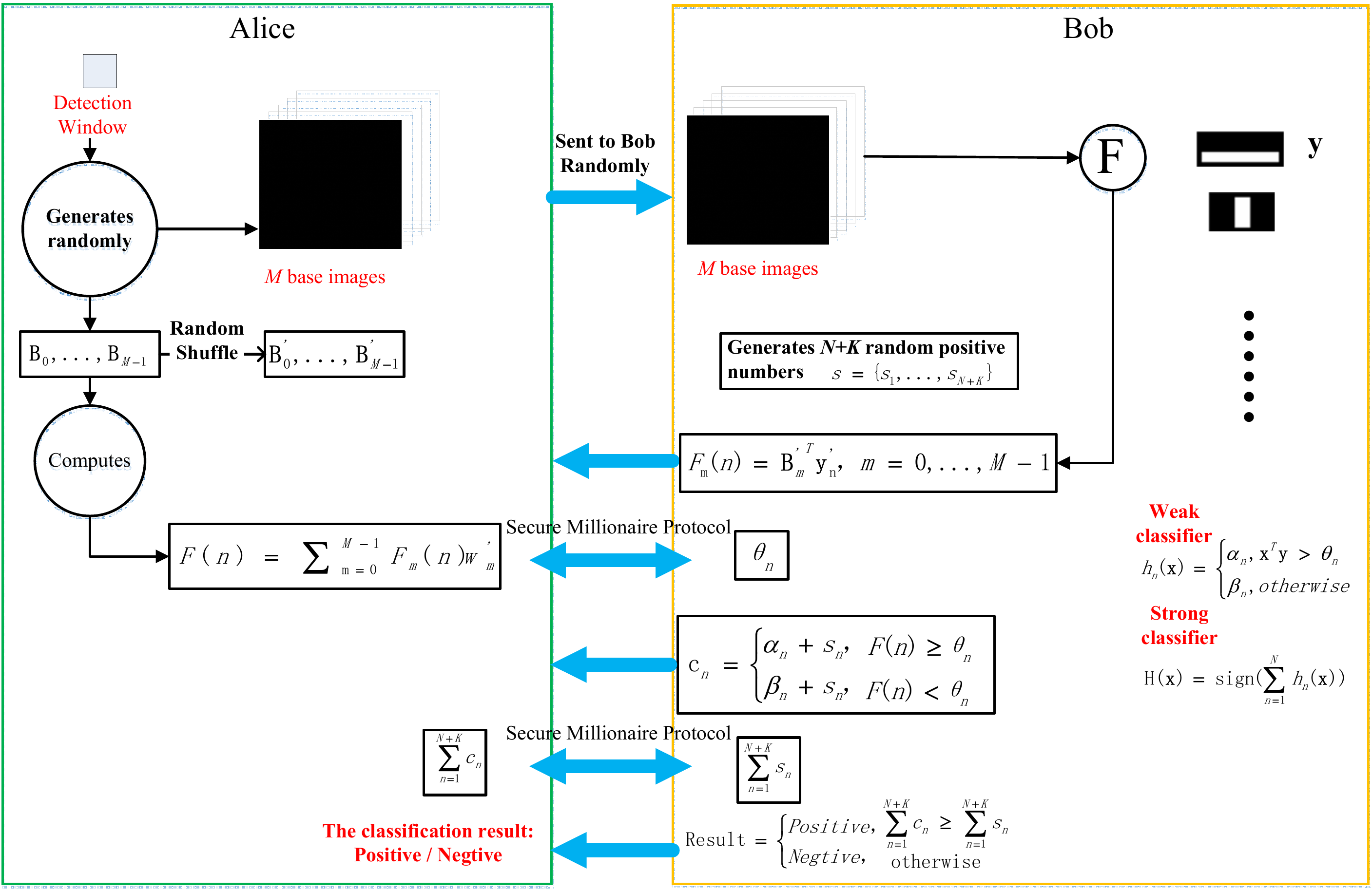}
\caption{The proposed secure object classifier. }
\label{fig:secure}
\end{figure*}

The core of our method is the secure object classifier protocol as is described in Algorithm \ref{alg:secure} and Figure \ref{fig:secure}. For secure object detection, Alice first divides the test image $\mathbf{X}$ into $Q$ detection windows $\{\mathbf{x}_1,\mathbf{x}_2,...\mathbf{x}_Q\}$. Then the detection windows are randomly sent to Bob as the inputs of the secure face classifier protocol one by one. Using the Algorithm \ref{alg:secure}, Alice and Bob know which detection windows are the target objects. Because the detection windows are randomly sent to Bob, only Alice learns the location of all the detected faces in the original image. Bob does not learn the contents including where the faces are in the image of Alice. Alice learns nothing about the parameters of the face detector of Bob.


\begin{algorithm}[htb]         
\caption{Secure Object Classifier Algorithm with RBI}             
\label{alg:secure}                  
\begin{algorithmic}[1]                
\REQUIRE ~~\\                       
    (1) Alice has input detection window $\mathbf{x} \in F^L$\\
    (2) Bob has a strong classifier of the form $H(\mathbf{x}) = \operatorname{sign}(\sum_{n=0}^{N-1}h_n(\mathbf{x}))$
\ENSURE ~~\\                         
    (1) Alice has the result $H(\mathbf{x})$ and nothing else\\
    (2) Bob learns nothing about the detection window $\mathbf{x}$

\end{algorithmic} 
\end{algorithm} 

The body of Algorithm \ref{alg:secure} is described as follows:

\begin{itemize}
\item (1): Alice factorizes the detection window $x$ into $M$ random base images $\mathbf{B}=\{\mathbf{B}_0, \mathbf{B}_1,..., \mathbf{B}_{M-1}\}$ with weight $w=\{w_0,w_1,...,w_{M-1}\}=\{0,1,...,M-1\}$ through Algorithm \ref{alg:RBI}.

\item (2): Alice randomly shuffles the weight $w$ to $w'$. The random base images $\mathbf{B}$ are permuted with the same order of that of $w'=\{w'_0,w'_1,...,w'_{M-1}\}$ to $\mathbf{B}'=\{\mathbf{B}'_0, \mathbf{B}'_1,..., \mathbf{B}'_{M-1}\}$, which is sent to Bob.

\item (3): In one cascade, Bob has $N$ weak classifiers with parameter vectors $\mathbf{y}=\{\mathbf{y}_0,\mathbf{y}_1,...\mathbf{y}_{N-1}\}$. Bob randomly add $K$ fake weak classifiers and set their parameters $\alpha$ and $\beta$ to zero. Bob randomly shuffles the $N+K$ true and fake weak classifiers to form $\mathbf{y}'=\{\mathbf{y}'_0,\mathbf{y}'_1,...\mathbf{y}'_{N+K-1}\}$. Then, Bob generates $N+K$ random positive numbers $s=\{s_0,s_2,...,s_{N+K-1}\}$. For each parameter vector $\mathbf{y}'_n \in \mathbf{y}'$. Bob and Alice repeat the following 3 steps.\\

\begin{itemize}

\item (3.1): Bob computes the feature responses for all the base image $\mathbf{B}'_m$ in $\mathbf{B}'$ by $F_m(n)={\mathbf{B}}_{m}^{'T}\mathbf{y}'_n, m=0,1,...,M-1$. All the $M$ responses of base images $\mathbf{B}'$ on each parameter vector $\mathbf{y}'_n$ are sent back to Alice.

\item (3.2): Alice computes the feature responses of the detection window $\mathbf{x}$ by $F(n) = \sum_{m=0}^{M-1}F_m(n)w'_{m}$.

\item (3.3): Alice and Bob use the secure Millionaire protocol \cite{Avidan2006a} to determine which number is larger: $F(n)$ or $\theta_n$. Bob send $\alpha_n+s_n$ or $\beta_n+s_n$ to Alice. Alice store it as $c_n$.

\end{itemize}

\item  (4): Alice and Bob use the secure Millionaire protocol \cite{Avidan2006a} to determine which number is larger: $\sum_{n=1}^{N+K}c_n$ or $\sum_{n=1}^{N+K}s_n$. If Alice has a larger number then x is positively classified, otherwise x is negatively classified.

\end{itemize}

\subsubsection{Security} The protocol protects the security of both parties. The protocol protects the contents of the image from Alice and the parameters of the face detector from Bob. We analyse the security of Algorithm \ref{alg:secure} in the following paragraph.

\begin{itemize}
 \item From Alice to Bob
 
 	\begin{itemize}
 	\item In step 2, Alice send randomly shuffled base images to Bob. Bob only knows the randomly generated base images and do not know the weight of each base image. The probability of guessing out the right permutation is $1/M!$. Even Bob guesses out the right permutation, he does not know the weight of each base image. Thus, it is almost impossible for Bob to recover the detection window of Alice.
 	\item In the 3th sub-step of step 3 and the step 4. Alice and Bob engage in secure Millionaire protocol \cite{Avidan2006a}. so Bob can learn nothing about Alice’s data.
 	\end{itemize}

 \item From Bob to Alice
 
 	\begin{itemize}
	\item In the 1st sub-step of step 3, Alice can not learn the number of the weak classifiers $N$ or the true filters from the received feature responses. The true filters are obfuscated by the fake filters. 
 	\item In the 3rd sub-step of step 3, Alice and Bob engage in a secure Millionaire protocol so Alice only learns if $F(n)>\theta_n$. She can not learn anything about the parameter $\theta_n$. Moreover, at the end of the Millionaire protocol Alice learns either $\alpha_n+s_n$ or $\beta_n+s_n$. In both cases, the real parameter ($\alpha_n$ or $\beta_n$) is obfuscated by the random number $s_n$.
	\item In step 4, Alice and Bob use the secure Millionaire protocol to determine which number is larger: $\sum_{n=1}^{N}c_n$ or $\sum_{n=1}^{N}s_n$. If Alice has a larger number then $x$ is positively classified, otherwise x is negatively classified.
 	
 	 \end{itemize}
 \item Multiple Cloud Servers
 	\begin{itemize}
 	\item The $M$ random base images can be also sent to multiple cloud server with the same object detector to increase security.
 	 \end{itemize}
 	 
\end{itemize}

\subsubsection{Complexity and Efficiency} The complexity of the protocol is $O(M(N+K)L)$, where $M$ is the number of the base images. $N$ and $K$ are the numbers of the true and fake weak classifiers, respectively. $L$ is the dimensionality of the detection window $x$.

Unlike the traditional Blind Vision \cite{Avidan2006a}, in which the OT operation is used extensively, the proposed method only use OT operation to compare 2 numbers. In the secure dot-product protocol, each pixel of each detection window uses a ${OT}_{1}^{256}$ operation, which needs $1$ RSA encryption and $256$ RSA decryption with 128-bit long encryption keys. We leverage our $M$ random images, whose computation is much faster than the RSA encryption and decryption operations.

In addition, in the traditional Blind Vision \cite{Avidan2006a}, they convert the integral-image representation to regular dot-product operation, a step that clearly slows down their implementation as they no longer take advantage of the integral-image representation. In our RBI based protocol, the integral-image representation is enabled again, which accelerates the computation obviously.

\section{Experiments}
We convert the Viola-Jones type object detector \cite{Viola2001,Viola2004} to our secure object detector. We implement our RBI based object detector using Microsoft Visual Studio 2012 and OpenCV 2.4.3/10. \footnote{http://opencv.org/} package for computer vision in a 64 bits Windows 7 operating system. The hardware configuration is 3.5GHz AMD A10 Pro-7800 R7 CPU with 12 compute Cores and 8GB Memory. 

The face detector is from the OpenCV 2.4.3 package and consists of a cascade of 22 rejectors, where each rejector is of the form presented in Eq. \ref{eq:strong}. The first rejector consists of 3 weak classifiers. The most complicated rejector consists of 213 weak classifiers. There is a total of 2135 weak classifiers. We also test the nose detector, the eye detector and the full body detector from OpenCV 2.4.10.



\subsection{The Detection Accuracy}


We test our secure face detector in 3 face detection datasets: The Face Detection Dataset (FDDB) \cite{fddbTech}, The Face96 Dataset \cite{Face96}, and The FEI Face Database \cite{Thoma2010}.

%
%

We randomly select 100 face images from each of the 3 datasets. The detection accuracy (88.46\%) of our secure face detector is the same as that of the OpenCV 2.4.3 face detector (88.46\%). 

The nose and the eye detectors are tested on the FDDB dataset \cite{fddbTech}. The full body detector is tested on the INRIA Person dataset \cite{Dalal2005}. We randomly select 100 images from each of the 2 datasets. The detection accuracy of our secure object detectors is the same as that of the OpenCV 2.4.10 nose, eye and full body detectors.


%
%
%
%

\subsection{Comparison with Other Methods} 
We compare our method with the  Viola-Jones method implemented by the OpenCV package and the method of the traditional Blind Vision \cite{Avidan2006a}. 50 test images with size of $100 \times 100$ are randomly selected from each of the 3 datasets. The average running time is shown in Table \ref{tb:TimeComp}. All the methods are running in client and server mode. For the Viola-Jones, Alice send the original image to Bob. Then, Bob runs the Viola-Jones method and return the detected windows to Alice. Our method is slower than the Viola-Jones method, which is running on plain images without protecting any privacy. According to the traditional Blind Vision method \cite{Avidan2006a}, the time-consuming OT operation is heavily used and the integral-image representation is disabled. Thus, they have to take a couple of hours to scan a single image, which is painfully slow. Although in our method, the only information that Bob learns is that how many faces are in the image of Alice, our cryptographic-free method is significantly faster than the previous work towards practical usage of blind vision applications.

In addition, we compare our method with the Viola-Jones method implemented by the OpenCV package and the method of the traditional Blind Vision \cite{Avidan2006a}. 50 test images with size of $100 \times 100$ are randomly selected from each of the FDDB and the INRIA Person datasets. The average running times are shown in the last 3 rows of Table \ref{tb:TimeComp}. All the methods are running in client and server mode.

\begin{table*}
\centering
\renewcommand{\multirowsetup}{\centering}  
\begin{tabular}{|c|c|c|c|c|c|} 
\hline  
\textbf{Dataset} & \textbf{Our} & \textbf{Our + Comm. Delays} & \textbf{VJ} \cite{Viola2004} & \textbf{VJ} \cite{Viola2004} \textbf{ + Comm. Delays} & \textbf{Blind Vision} \cite{Avidan2006a} \\
\hline
FDDB-face  & 143.852s & 380.992s & 0.380s & 0.843s & 
\multirow{3}{2cm}{A couple of hours \cite{Avidan2006b}} \\
Face96  & 173.471s & 477.635s &  0.358s & 0.809s & \\ 
FEI     & 152.701s & 414.522s &  0.363s & 0.827s & \\ 
\hline 
\hline
FDDB-nose  & 113.398s & 294.845s & 0.372s & 0.836s &
\multirow{3}{2cm}{A couple of hours \cite{Avidan2006b}} \\
FDDB-eye  & 85.754s & 240.111s &  0.496s & 0.912s &\\ 
INRIA Person   & 80.204s & 224.571s & 0.333s & 0.771s  &\\ 
\hline 
\end{tabular} 
\caption{Average running time comparison with the Viola-Jones method \cite{Viola2004} and the Blind Vision method \cite{Avidan2006a}  on the FDDB \cite{fddbTech}, the Face96 \cite{Face96} and the FEI  \cite{Thoma2010} datasets. In the second and fourth columns, we simulate Alice and Bob on one PC without communication delays. The third and fifth columns report the time costs in a private cloud environment.
}
\label{tb:TimeComp}
\end{table*}

\section{Conclusions and Discussions}
We propose a novel random base image representation (RBI) for efficient object detection applications. The traditional blind vision method applies secure multi-party techniques to vision algorithm. Their method reveals no information to either party at the expanse of heavy computation load. Our method is an attempt towards cryptographic-free. Alice learns nothing about the parameters of the face detector of Bob. Bob does not know the contents of the image of Alice. The only information may be leaked is that Bob have a probability $1/M!$ to guess out the right permutation of the base images. This is just a theoretical event. Even Bob guesses out the right permutation, he does not know the weight of each base image.  Thus it is almost impossible for Bob to learn the information of the detection window of Alice. Because the heaviest cost of OT operation in the secure dot-production of \cite{Avidan2006a} is avoided by our RBI based dot-production, the Millionaire version protocol of ours need much less time than the traditional blind vision protocol does.

There are several extensions to this work. First is the need to accelerate the secure blind vision to practical use, i.e. to reduce the time cost to near that of the vision algorithm without security consideration. Second is to make both the training and the test blind. This will make the client users to upload more visual data to the cloud without worrying about the privacy leakage.

\bibliographystyle{IEEEbib}
\bibliography{camera-ready_icme2017template}

\end{document}